\documentclass{article}

    \PassOptionsToPackage{numbers}{natbib}
 \usepackage[preprint]{neurips_2026}
\workshoptitle{Sim2Science}

\usepackage[utf8]{inputenc} 
\usepackage[T1]{fontenc}    
\usepackage{hyperref}       
\usepackage{url}            
\usepackage{booktabs}       
\usepackage{amsfonts}       
\usepackage{nicefrac}       
\usepackage{microtype}      
\usepackage{xcolor}         
\usepackage{wrapfig}
\usepackage{siunitx}
\usepackage{graphicx}
\usepackage{adjustbox}
\usepackage{algorithm}
\usepackage{algorithmic}
\usepackage{amsmath}
\usepackage{fontawesome5}
\title{Probabilistic Simulation of Gas-Liquid Interfaces with Coupled Flow Matching on Spherical Harmonics}

\definecolor{darkblue}{rgb}{0.0, 0.0, 0.5}
\definecolor{darkred}{rgb}{0.5, 0.0, 0.0}
\hypersetup{
    colorlinks=true,
    citecolor=darkblue, 
    linkcolor=darkred, 
    urlcolor=darkblue   
}

\author{%
Kiet Bennema ten Brinke$^{1}$ \quad
Rachna Ramesh$^{1}$ \quad
Douwe Orij$^{2}$ \quad
Ivo Roghair$^{2}$ \quad \\
\textbf{Vlado Menkovski}$^{1}$ \\[2mm]
{\small
$^{1}$ Department of Mathematics and Computer Science,
Eindhoven University of Technology} \\
\small{
$^{2}$ Department of Chemical Engineering and Chemistry,
Eindhoven University of Technology} \\[1mm]
{\small
\texttt{\{k.bennema.ten.brinke, r.ramesh1, d.r.orij,  i.roghair, v.menkovski\}@tue.nl} 
}
}

\begin{document}

\maketitle

\begin{abstract}

Bubbly flows involve strongly coupled bubble motion, deformation, and hydrodynamic interactions, making high-fidelity simulation computationally expensive. We propose a neural surrogate that jointly predicts bubble trajectories and interface deformation in a compact, shared representation, replacing costly mesh-resolved simulation with fast generative inference. Because two-phase flow dynamics is nonlinear and chaotic, we treat forecasting as a conditional generative problem rather than deterministic regression, predicting entire future windows in a single pass. We train and evaluate the surrogate on BubbleSH, a Front-Tracking dataset of rising deformable bubble swarms, assessing performance using trajectory, shape, and distributional metrics over kinematic, morphological, and interaction-based properties. The surrogate reproduces key flow statistics while running several orders of magnitude faster than the underlying DNS solver. \href{https://anonymous.4open.science/r/BubbleSim-Sim2Sci}{\texttt{Link to Code} \ \faGithub}.
\end{abstract}

\vspace{-1em}
\begin{figure}[H]
\adjustbox{trim=0.1cm 0.0cm 0cm 0.0cm}{
    \centering
    \includegraphics[width=0.98\textwidth]{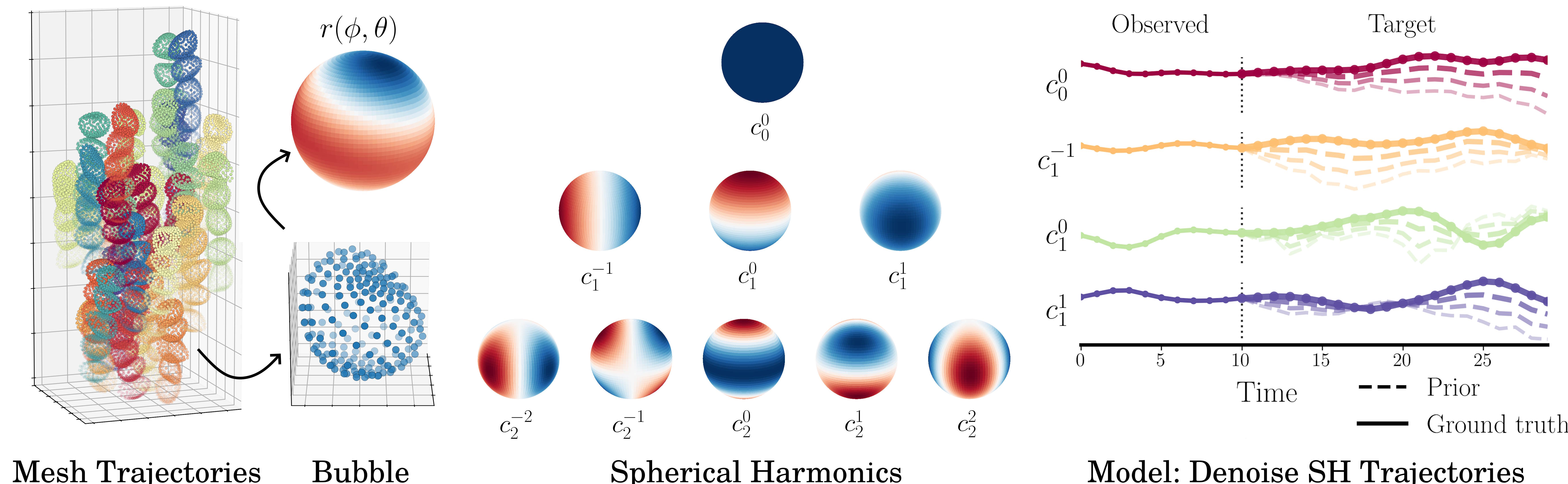}
    }
\end{figure}
\vspace{-2em}

\section{Introduction}
Bubbly two-phase flows underlie a wide range of industrial and natural processes, from chemical reactors and water treatment to CO$_2$ sequestration and volcanic degassing, where the collective motion of deformable gas-liquid interfaces governs mass and heat transfer at the system scale \cite{prosperetti2009computational}. Predicting this behavior requires resolving both the trajectories of individual bubbles and their time-varying shapes, which are coupled through the surrounding liquid via pressure, drag, and wake interactions \cite{Roghair2016}.

High-fidelity Direct Numerical Simulation (DNS) methods such as Front-Tracking \cite{Unverdi1992, SintAnnaland2006, Roghair2016} and Volume-of-Fluid \cite{Hirt1981, SintAnnaland2005} solve the incompressible Navier-Stokes equations while explicitly resolving the gas-liquid interface on a high-resolution mesh. These methods are considered the gold standard for accuracy but scale poorly as interface reconstruction and frequent remeshing dominate the computational cost, making swarms of more than a few dozen bubbles prohibitively expensive to simulate for any appreciable physical duration \cite{Roghair2016}. Euler-Lagrange approaches trade this fidelity for speed by representing bubbles as point particles with empirical closure laws, enabling large-scale simulation at the cost of discarding interface-resolved information entirely, deformation, breakup, and near-field interaction effects are lost \cite{LAU20113309}. This leaves a gap between two regimes, namely mesh-resolved simulators that are accurate but too slow to scale, and reduced-order simulators that scale but cannot represent deforming interfaces. 

We address this gap with a data-driven neural surrogate that jointly forecasts bubble trajectories and interface deformation in a compact, shared representation. Each bubble's interface is encoded with spherical harmonics, exploiting its star-shaped geometry to replace dense meshes with a small set of shape coefficients. Because bubble dynamics are chaotic, we treat forecasting as a generative rather than deterministic problem, and design the model to limit compounding error that typically plagues long-horizon rollouts.

The surrogate is evaluated on BubbleSH \cite{ramesh2026bubbleshdatasetrisingbubbles}, a Front-Tracking dataset of rising bubble swarms, using point-based and distributional metrics spanning kinematic, morphological, and interaction statistics. Our surrogate closely reproduces these statistics while running several orders of magnitude faster than the DNS solver it emulates, making swarm-scale, interface-resolved simulation more practical.


\section{Preliminaries} \label{sec:preliminaries}

\textbf{Problem Formulation}
We consider the problem of probabilistically forecasting the evolution of interacting gas--liquid interfaces from a finite history of their positions and shapes. Let $B$ denote the number of bubbles and represent the state at time $t$ by $x_t = \left\{\mathbf{p}_t^{(b)}, \mathbf{c}_t^{(b)}\right\}_{b=1}^{B}$
where $\mathbf{p}_t^{(b)} \in \mathbb{R}^3$ is the centroid position of bubble $b$ and $\mathbf{c}_t^{(b)} \in \mathbb{R}^{(L+1)^2}$ contains its spherical-harmonic shape coefficients. Given a conditioning window $X_{0:T_c} = (x_0,\ldots,x_{T_c})$, the objective is to learn the conditional distribution of future states $p\left(X_{T_c+1:T} \mid X_{0:T_c}\right)$.
Forecasting is formulated as conditional generative modeling, where samples are drawn from a prior $p_0$ and a mapping is learned to the target future-state distribution $p_1(X_{T_c+1:T})$.

\textbf{Spherical Harmonics Representation}
Any square-integrable function can be represented as a unique combination of spherical harmonics basis functions, analogous to the Fourier transform but on the spherical domain instead of the frequency domain. We focus on sphere-like objects whose radius can be expressed as a function on the unit sphere $r(\theta, \phi)$, where $\theta \in [0, 2\pi]$ is the azimuthal angle and $\phi \in [0, \pi]$ is the polar angle. This function is expanded in the spherical harmonic basis $Y^m_{\ell}(\theta, \phi)$ \cite{Brechbuhler1992}:
\begin{equation} \label{sph}
    r(\theta, \phi) = \sum_{\ell=0}^{L} \sum_{m=-\ell}^{\ell} c_\ell^m \, Re(Y^m_{\ell}(\theta, \phi)), \ \ \ Y_{\ell}^{m}(\theta,\phi)
    = 
    \sqrt{\frac{(2\ell+1)(\ell-m)!}{4\pi(\ell+m)!}}
    \,P_{\ell}^{m}(\cos\phi)\,e^{im\theta}
\end{equation}
where $c_\ell^m \in \mathbb{R}$ are the real spherical harmonic coefficients, $P_{\ell}^{m}$ denotes the associated Legendre polynomial with the Condon--Shortley phase convention and $L$ is the
truncation degree. Truncating at degree $L$ yields a fixed-size representation
$\mathbf{c} \in \mathbb{R}^{(L+1)^2}$, which for smooth, star-shaped geometries like bubbles concentrates most of the shape's energy in the low-order coefficients, providing a more compact and efficient representation than standard meshes \cite{yu3DBubblesExperimentalDataset2026}.

\textbf{Generative Framework}
Flow matching \citep{lipman2023, Liu2022FlowSA, albergo2024stochastic} (i.e. Rectified flow, Stochastic Interpolants) learns a time-dependent vector field
$v_\theta(\mathbf{x}_\tau, \tau)$ that transports samples from a prior distribution
$p_0$ to a target data distribution $p_1$ along a probability path
$p_t(\mathbf{x})$. Given a coupling $\pi$ between prior sample $\mathbf{x}_0 \sim p_0$
and data sample $\mathbf{x}_1 \sim p_1$, a conditional path is typically defined as
$ \mathbf{x}_\tau = (1-\tau)\,\mathbf{x}_0 + \tau\,\mathbf{x}_1, \ \tau \in [0,1]$
with corresponding conditional target velocity $\mathbf{u}_\tau = \mathbf{x}_1 - \mathbf{x}_0$.
The model is trained by regressing $v_\theta$ onto this target velocity via the
conditional flow matching objective:
\begin{equation} \label{fmloss}
    \mathcal{L}_{\text{CFM}}(\theta) = \mathbb{E}_{\tau\sim \mathcal{U}[0,1], (\mathbf{x}_0, \mathbf{x}_1)\sim \pi}
    \left[ \left\| v_\theta(\mathbf{x}_\tau, \tau) - \mathbf{u}_\tau \right\|^2 \right].
\end{equation}
At inference time, samples are generated by integrating $v_\theta$ from $\tau=0$ to
$\tau=1$ starting from a draw of the prior $p_0$.

\textbf{Data-Dependent Couplings} The coupling $(\mathbf{x}_0,\mathbf{x}_1)\sim \pi$ used to draw training and inference samples is most often defined to be independent of targets, $\pi=p_0(\mathbf{x}_0)p_1(\mathbf{x}_1)$. We define $\pi$ to be dependent on the conditioning window $X_{0:T_c}$ as we are modeling the conditional distribution $p(X_{T_c+1:T} \mid X_{0:T_c})$, using the notion of a data-dependent coupling, $\pi(\mathbf{x}_0,\mathbf{x}_1)=\pi(\mathbf{x}_0 \mid \mathbf{x}_1)p_1(\mathbf{x}_1)$ \cite{albergo2024stochastic}. Priors $p_0$ that decrease the required transport cost of moving probability mass to $p_1$ are shown to increase model efficiency \cite{brinke2026, tong2023improving}.

\section{Methodology} \label{sec:methodology}

\textbf{Data Representation and Features} Positions are represented in the cylindrical coordinate system, $\mathbf{p}_t^{(b)}=(r,\phi,z)$, to align with the rotational symmetry along the vertical axis due to buoyancy. The interfaces defined on $r(\theta, \phi)$ are fitted on the spherical harmonic basis functions $Y_l^m(\theta, \phi)$ \eqref{sph} with coefficients $\mathbf{c}$, determined by a least squares linear solver on $\mathbf{Yc}=\mathbf{r}$, where $\mathbf{Y} \in \mathbb{R}^{N\times M}$ are the $M$ basis functions evaluated at all $N$ $(\theta, \phi)$ vertex pairs, $\mathbf{c} \in \mathbb{R}^{M \times 1}$ and $\mathbf{r} \in \mathbb{R}^{N \times 1}$. Our reference DNS model uses on average $3.9\cdot10^4$ floats per bubble shape, while we use $M=36$ SH coefficient floats, totaling a compression ratio of $1083 \times$ with only minor loss in reconstruction quality \cite{yu3DBubblesExperimentalDataset2026}.  A radius graph is constructed for each $x_t$ with node features consisting of velocities $\Delta \mathbf{p}^{(b)}_t$, and accelerations $\Delta^2 \mathbf{p}^{(b)}_t$ and coefficients $\mathbf{c}^{(b)}_t$, $\Delta \mathbf{c}^{(b)}_t$, along with edge features based on distance and node feature differences.

\textbf{Prior}
Following the data-dependent coupling formulation, we construct the prior sample
$\mathbf{x}_{\tau=0}$ from a per-bubble random walk fit to the conditioning window
$X_{0:T_c}$, applied identically to the cylindrical position and
shape-coefficient channels. For each bubble, we estimate a mean step
$\mu$ and step variance $\sigma^2$ from the finite differences of the
observed prefix, then extrapolate forward according to
$
    x_t
    =
    x_{t-1}
    +
    \mu
    +
    \sigma \odot s \odot z_t,
    \ \
    z_t \sim \mathcal{N}(0,I),
    \ \ 
    t \in \{T_c,\ldots,T-1\}$
with noise-level hyperparameter $s$. The observed prefix
is kept exact, i.e.,
$
    X_{\tau=0}^{0:T_c} = X_{\tau=1}^{0:T_c},
$
with only the unobserved future $X_{\tau=0}^{T_c:T}$ drawn from the random walk. This construction
reduces the transport cost that the vector field $v_\theta$ must learn
relative to a fixed Gaussian prior, since the random walk already captures
each bubble's low-order motion and deformation trends \cite{brinke2026}.

\textbf{Architecture}
The vector field $v_\theta$ alternates spatial and temporal layers, following \cite{brinke2026}. The spatial component is a  Neural Message Passing GNN \cite{pmlr-v70-gilmer17a}, where messages are computed per edge from sender and receiver hidden states and edge
features, aggregated by summation, and used to update
node hidden states with a residual MLP block. The temporal component is a
1D-convolutional U-Net \cite{dhariwal2021diffusionmodelsbeatgans} operating along the trajectory axis, conditioned on the
flow-matching time $\tau$ via a sinusoidal embedding. An additional physical frame
index sinusoidal embedding is added to the node features.
Spatial and temporal layers are interleaved across the network depth and velocity predictions are decoded at every temporal layer and added residually. We
attach a separate MLP decoder per spherical-harmonic band $\ell$, mapping the
final hidden state to the predicted velocity of each band's coefficients
independently, as the spherical harmonics form an orthonormal basis. 

\textbf{Training and Generation}
We train the vector field $v_\theta$ to model the conditional distribution
$
    p\!\left(
        X_{T_c+1:T}
        \,\middle|\,
        X_{0:T_c}
    \right)
$
jointly over positions and shape coefficients, using time-bundled flow
matching. Rather than predicting the state trajectories directly, we set the targets as the change between consecutive states, $\Delta \mathbf{p}_t^{(b)}$ for the cylindrical position and $\Delta \mathbf{c}_t^{(b)}$ for the shape coefficients.
Instead of stepping frame-by-frame, $v_\theta$ denoises the
time-differences of the entire future window $X_{T_c+1:T}$ in a single forward
pass, reducing the compounding-error failure mode of autoregressive
rollouts. We supervise the two channels with
separate losses, $\mathcal{L}_{\mathrm{pos}}$ and
$\mathcal{L}_{\mathrm{SH}}$, each defined as Eq.~\eqref{fmloss} restricted to that channel, with total loss $
    \mathcal{L}
    =
    \mathcal{L}_{\mathrm{pos}}
    +
    \lambda \mathcal{L}_{\mathrm{SH}}$. At inference, we integrate $v_\theta$ from the random walk prior $\mathbf{x}_{\tau=0}$ to
$\tau=1$ using the Euler scheme described in Algorithm \ref{algo2}, recovering the
full future trajectory and interface deformation in $10$ denoising steps. See Appendix \ref{app:model} for more details about the model.

\section{Experiments} \label{sec:experiments}

\definecolor{posColor}{RGB}{30, 58, 105}    
\definecolor{shapeColor}{RGB}{20, 85, 70}    
\definecolor{distColor}{RGB}{110, 40, 55}   

\begin{table}[t!]
\vspace{-3em}
\caption{Experimental study comparing the prior baselines and our generative model with different priors, formulations and architectures expressed in point-based and distributional metrics.}
\centering
\small
\setlength{\tabcolsep}{4pt}
\adjustbox{trim=0.5cm 0.5cm 0cm 0.0cm}{
\begin{tabular}{l cccccc c}
\toprule
 & \multicolumn{2}{c}{\textbf{\textcolor{posColor}{Positional Errors}}} 
 & \multicolumn{2}{c}{\textbf{\textcolor{shapeColor}{Deformation Errors}}} 
 & \multicolumn{3}{c}{\textbf{\textcolor{distColor}{Distributional Alignment ($\mathcal{W}_{IQR}$})}} \\
\cmidrule(lr){2-3} \cmidrule(lr){4-5} \cmidrule(lr){6-8}

 & \textcolor{posColor}{R-ADE $\downarrow$} 
 & \textcolor{posColor}{R-FDE $\downarrow$} 
 & \textcolor{shapeColor}{IoU $\uparrow$} 
 & \textcolor{shapeColor}{CD $\downarrow$} 
 & \textcolor{distColor}{{\footnotesize Kinematics} $\downarrow$} 
 & \textcolor{distColor}{{\footnotesize Morphology} $\downarrow$} 
 & \textcolor{distColor}{{\footnotesize Interaction} $\downarrow$} \\
\midrule

{\tiny i } \ \  \textcolor{gray}{Gaussian prior} & \textcolor{gray}{0.617} & \textcolor{gray}{1.027} & \textcolor{gray}{0.708} & \textcolor{gray}{0.605} & \textcolor{gray}{2.529} & \ \textcolor{gray}{0.634} & \textcolor{gray}{0.065} \\ 
{\tiny ii } \   \textcolor{gray}{Informed prior} & \textcolor{gray}{0.267} & \textcolor{gray}{0.521} & \textcolor{gray}{0.569} & \textcolor{gray}{1.023} & \textcolor{gray}{0.209} & \textcolor{gray}{0.462} & \ \textcolor{gray}{0.096} \vspace{0.3em} \\ 

{\tiny iii} \ Model \footnotesize{$(L=5)$} & 0.176 & 0.309 & \textit{0.751} & 0.516 & 0.164 & \textcolor{distColor}{\textit{0.229}} & 0.050 \\ 
{\tiny iv} \ \ \ \ {\tiny with} \textit{Gaussian prior} & \textcolor{posColor}{\textit{0.160}} & \textcolor{posColor}{\textbf{0.284}} & 0.750 & 0.512 & 0.258 & 0.357 & 0.040 \\ 

{\tiny v} \ \ \ \ \ {\tiny with} \textit{Determinism} & 0.173 & 0.300 & \textcolor{shapeColor}{\textbf{0.762}} & \textcolor{shapeColor}{\textbf{0.489}} & 0.311 & 0.495 & \textcolor{distColor}{\textbf{0.034}} \\ 
{\tiny vi} \ \ \ \ {\tiny with} \textit{Independence} & 0.189 & 0.327 & 0.748 & 0.528 & \textcolor{distColor}{\textbf{0.140}} & \textcolor{distColor}{\textbf{0.198}} & 0.047 \\ 

{\tiny vii } \ \ \ {\tiny with} \textit{Transformer} & 0.186 & 0.332 & 0.738 & 0.550 & 0.164 & 0.243 & \ 0.057 \vspace{0.3em} \\ 
{\tiny viii} \  Model \footnotesize{$(L=3)$} & \textcolor{posColor}{\textbf{0.158}} & \textcolor{posColor}{\textit{0.285}} & \textcolor{shapeColor}{\textit{0.755}} & \textcolor{shapeColor}{\textit{0.504}} & \textcolor{distColor}{\textit{0.163}} & 0.414 & \textcolor{distColor}{\textit{0.036}} \\ 
\bottomrule
\end{tabular} }
\label{results} 
\end{table}

\textbf{Dataset} The neural surrogate is trained and tested on Front Tracking simulator data from the dataset called BubbleSH \cite{ramesh2026bubbleshdatasetrisingbubbles}. The dataset contains centroid and interface trajectories of $B=32$ bubbles governed by the incompressible Navier-Stokes equations simulated at $10^{3}$ Hz, using configurations of bubble sizes $d\in\{4,5,6\}$ mm and gas volume fractions $\varepsilon\in\{5,10,...,40\}$. The bubble interfaces are represented with spherical harmonics coefficients $\mathbf{c}\in \mathrm{R}^{225}, L=14$ and fixed length trajectory windows of 60 ms and $T=30$ frames are extracted using a sliding-window approach. For each window, the first 20 ms ($T_c=10$) are used as the observed conditioning prefix, and the remaining portion is the prediction target. The processed data contains $2094$ samples of 30 frames in total. 

\begin{wrapfigure}[17]{R}{0.35\textwidth}
\adjustbox{trim=-0.2cm 0.0cm 0.0cm 0.2cm}{
    \includegraphics[width=0.95\linewidth]{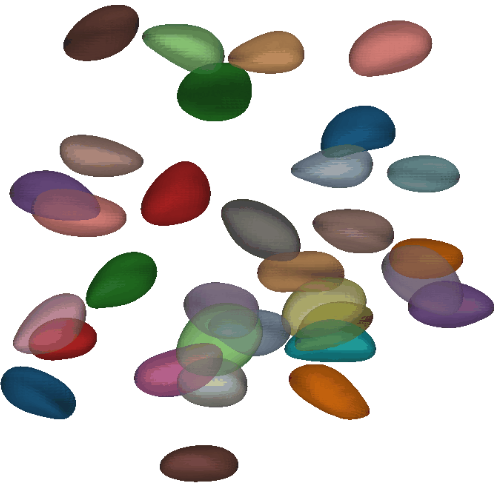}}
    \caption{Snapshot of 32 bubble meshes from the BubbleSH dataset}
    \label{front}
\end{wrapfigure}

\begin{figure} 
\adjustbox{trim=0.7cm 0.6cm 0cm 1.3cm}{
    \includegraphics[width=1.04\textwidth]{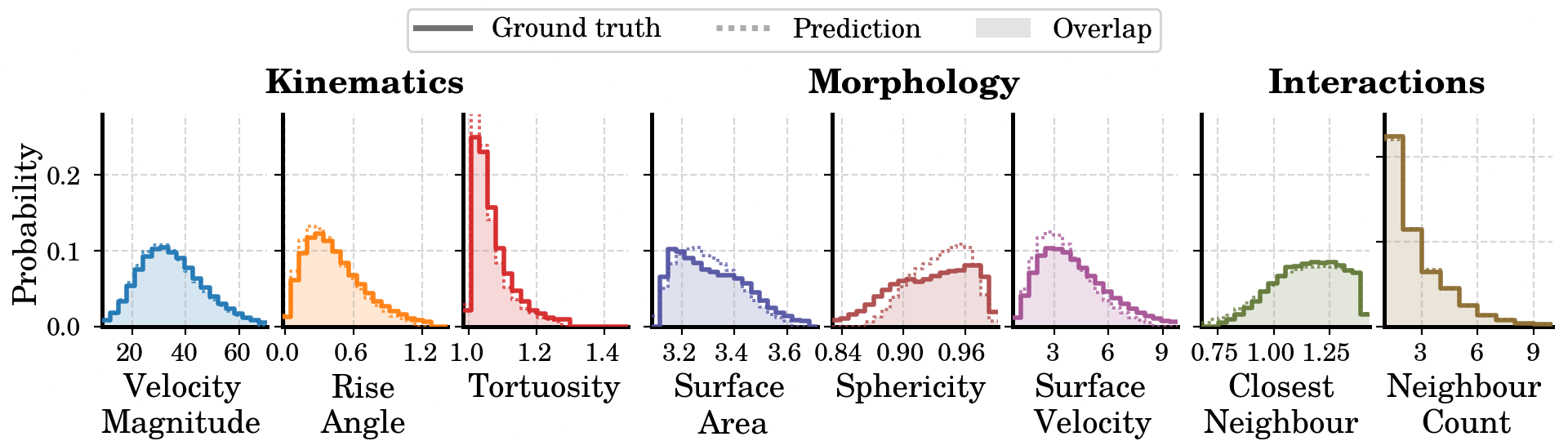}}
    \caption{Probability mass functions of marginal kinematic, shape and interaction properties calculated from predicted and ground truth target windows from experiment \hyperref[results]{(\textit{iii})}. Unit of distance is bubble diameter, unit of time is seconds. See Appendix \ref{app:eval-details} for details.}
    \label{comp}
    \vspace{-1.5em}
\end{figure}

\textbf{Metrics} Accuracy is measured as point-wise errors on positional and shape data. The Relative Average Displacement Error, measuring the mean ratio of displacement error to arc length, is used on positions for a dimensionless and interpretable metric, defined as R-ADE $=\langle ||\mathbf{x}_i(t)-\hat{\mathbf{x}}_i(t)|| \ / \ \sum_{t=0}^T ||  \dot{\mathbf{x}}_i(t)|| \rangle$, where $\langle \cdot \rangle$ denotes the mean. It is accompanied by the Relative Final Displacement error. To evaluate deformation errors over time, mean Intersection over Union (IoU) and the mean Chamfer Distance (CD) is used, CD $=\langle CD(r_i(t),\hat{r}_i(t))\rangle$ with $CD(A,B)=\frac{1}{|A|}\sum_{a\in A}min_{b\in B}||a-b||+\frac{1}{|B|}\sum_{b\in B}min_{a\in A}||a-b||$ for bubble point clouds $A$ and $B$ \cite{DBLP:journals/corr/abs-2111-12702}. We also analyze statistical alignment in the form of IQR-normalized Wasserstein-1 distance, as $\mathcal{W}_{IQR}(p,q)= \mathcal{W}_1(p,q) /\langle IQR(p)+IQR(q) \rangle $, over marginal kinematic, morphological and interaction-based quantities from \cite{ramesh2026bubbleshdatasetrisingbubbles}, visualized in Figure \ref{comp}. 

\textbf{Setup} We compare two parameter-free priors as baselines and six model configurations: \hyperref[results]{\textit{(i)}} a baseline where the generated trajectory is sampled from a isotropic Gaussian $\mathcal{N}(x_{T_c},\sigma^2)$ centered at the last observed frame or \hyperref[results]{\textit{(ii)}} a random walk conditioned on the observed velocities;
\hyperref[results]{\textit{(iii)}} the presented model using $L=5,(5+1)^2=36$ SH coefficients; \hyperref[results]{\textit{(iv)}} the model using $\mathcal{N}(x_{T_c},\sigma^2)$ as prior; \hyperref[results]{\textit{(v)}} the model using deterministic regression instead of flow matching; \hyperref[results]{\textit{(vi)}} a variant where the bubbles are modeled as being conditionally independent by removing the spatial interaction layers; \hyperref[results]{\textit{(vii)}} the presented model with both the spatial and temporal layers replaced by Diffusion Transformers w/ RoPE + parallel self-attention, code from \cite{lamslide}; \hyperref[results]{\textit{(viii)}} the model using a coarser shape representation of $L=3,(3+1)^2=16$ SH coefficients. 

\textbf{Results} Both non-parameterized priors perform far worse than any trained configuration across every metric as expected, confirming that the denoising models  themselves, not merely the prior, are responsible for the predictive accuracy. The isotropic Gaussian baseline \hyperref[results]{\textit{(i)}} in particular yields an implausible Kinematics score. The full model \hyperref[results]{\textit{(iii)}} achieves a good overall balance between point-based accuracy and distributional realism. Replacing the informed prior with a Gaussian centered at $x_{T_c}$ \hyperref[results]{\textit{(iv)}} improves displacement metrics and Interaction alignment, but degrades Kinematics and Morphology alignment, while leaving deformation error (CD) relatively unchanged. Replacing flow matching with deterministic regression \hyperref[results]{\textit{(v)}} yields the best IoU and CD but worst Kinematics and Morphology alignment, indicating that mode-averaging primarily harms the marginal kinematic and shape statistics rather than the point-based ones.  Removing the spatial interaction layers \hyperref[results]{\textit{(vi)}} causes a modest increase in displacement errors, while improving Kinematics and Morphology alignment, suggesting that at the density and timescale present in our processed data, near-bubble interactions have limited impact on these two metrics. The transformer-based architecture \hyperref[results]{\textit{(vii)}} underperforms on nearly every metric while scaling worse with window length. Finally, coarsening the shape representation to $L=3$ yields the best positional errors since more of the model's parameter capacity is available for positional prediction, but morphology alignment is substantially worsened. More analysis and evaluation details can be found in Appendix \ref{app:eval}. Beyond accuracy, our neural surrogate achieves four to five orders of magnitude speedup over the Front Tracking simulator, noted in Table \ref{table:speed}.

Additionally, we plot probability mass functions of the noted kinematic, morphological and interaction-based properties from experiment \hyperref[results]{\textit{(iii)}} in Figure \ref{comp}. The kinematic and interaction-based histograms align well with the ground truth, however the model shows some bias towards predicting lower tortuosity trajectories and shapes with higher sphericity.

\begin{wraptable}[10]{R}{0.35\textwidth}
\vspace{-1.2em}
\centering
\footnotesize
\setlength{\tabcolsep}{4pt}
\caption{Comparison of required time to run 0.1s of simulation.}
\label{table:speed}
\vspace{0.7em}
\begin{tabular}{lccc}
    \toprule
     & $\varepsilon$ & Time& Speedup \\
    \midrule
    DNS & $5$ & $2.6$ days & $1\times$ \\
    DNS & $40$  & $0.6$ days & \footnotesize $1\times$ \\
    Surrogate & \footnotesize$5$ & $0.85$ sec  & \footnotesize$260$,$000\times$ \\
    Surrogate & \footnotesize$40$ & $0.85$ sec  & \footnotesize$58$,$000\times$ \\
    \bottomrule
\end{tabular}
\end{wraptable}


\section{Related Work}

\textbf{Neural Surrogates and Shape Representations for Hydrodynamics} 
Learning-based surrogates provide efficient alternatives or complements to conventional solvers, ranging from PINNs~\cite{PINN} to graph-based simulators such as GNS~\cite{GNSSanchez_Gonzalez} and MeshGraphNets~\cite{MeshGraphNets}. They have been applied to learned simulation of fluid and other physical systems~\cite{GNSSanchez_Gonzalez}, ML-accelerated CFD~\cite{kochkov}, and large-scale forecasting~\cite{fourcastnet}. Multiphase-flow benchmarks have also emerged, including BubbleML~\cite{BubbleML} for boiling and phase change and MPF-Bench~\cite{MPFBench} for Lattice-Boltzmann simulations of droplets and rising bubbles. In parallel, spherical harmonics provide compact representations of closed 3D surfaces~\cite{nortjeSphericalHarmonicsSurface2015a,shenLargeScaleModelingParametric2006a} and have been used to characterize bubble deformation and oscillation modes~\cite{aganinDynamicsGasBubbles2022,cattaneoShapeModesJet2025,inserraMathematicalModellingAcoustic2025}. The experimental 3DBubbles dataset~\cite{yu3DBubblesExperimentalDataset2026} further demonstrates their utility for bubble-shape representation, motivating their use here as a time-resolved state representation for interacting bubble swarms.

\textbf{Probabilistic Trajectory Models}
Probabilistic trajectory forecasting has evolved from latent-variable and autoregressive sequence models, including CVAE- and GAN-based approaches, which represent multimodal futures but generate them through recurrent or sequential decoding \cite{Xu_2022, amirian2019socialways, salzmann2021trajectrondynamicallyfeasibletrajectoryforecasting}. More recently, diffusion models have enabled direct modeling of joint future-trajectory distributions \cite{Mao_2023_CVPR, Jiang2023MotionDiffuserCM}. Diffusion-based models for geometric dynamics further incorporate spatial symmetries and structured priors \cite{GeoTDM}. In contrast to iterative score-based denoising, flow matching trajectory approaches can use non-diffusive probability paths \cite{ye2024efficient,lamslide}, while stochastic interpolants provide a broader framework encompassing both flow- and diffusion-based generation \cite{chen2026flowdas}. Most closely related to our formulation, STFlow combines flow matching with a conditioning-dependent random-walk prior and spatio-temporal networks to reduce transport cost for geometric trajectory simulation \cite{brinke2026}. Our approach builds on this perspective but targets coupled multiphase dynamics.

\section{Conclusion} \label{sec:conclusion}
Simulating deformable bubble swarms traditionally forces a trade-off between accurate but slow mesh-resolved DNS and fast but interface-agnostic Euler-Lagrange models. We close this gap with a neural surrogate that jointly forecasts bubble positions and gas-liquid interfaces, representing 3D shapes with a compact spherical-harmonics expansion and forecasting them with a time-bundled flow matching model conditioned on an informed random-walk prior. Evaluated on Front-Tracking simulations, our surrogate reproduces bubble kinematics, morphology, and interaction statistics while running orders of magnitude faster than the DNS solver it emulates, demonstrating deep generative models as a practical, interface-resolved complement to classical hydrodynamics simulators.

\textbf{Limitations and Future Work} Our implementation is limited to fixed-length trajectories per dataset. Long rollout stability and bifurcations have not been evaluated. More efficient neural architectures for learning the spherical harmonics representation equivariantly can be designed. 

\newpage

\bibliographystyle{plainnat}
\bibliography{bibliography}

\newpage




\appendix

\section{Model Details} \label{app:model}

\subsection{Model Training and Inference Procedures} \label{appendix:algos}
The training and inference algorithms, inspired from \cite{brinke2026} are denoted below in Algorithm \ref{algo1} and Algorithm \ref{algo2}.

\begin{algorithm}[]

\caption{Conditional Flow Matching Training}
\label{algo1}
\centering
\begin{algorithmic}

\REQUIRE Train Dataset $p_1$, Model $v_\theta(\mathbf{x}, \mathbf{h}, \mathcal{E}, \tau)$
\WHILE{Training}
\STATE $\mathbf{x}_1 \sim p_1, \ \tau \sim \mathcal{U}[0,1]$ \hfill \# Sample batch $\mathbf{x}_1$ and denoising level $\tau$
\STATE $\mathbf{x}_0^{0:c} \gets \mathbf{x}_1^{0:c}$, \ $\mathbf{x}_0^{c:T} \gets \zeta(\mathbf{x}_1^{0:c})$
 \hfill \#  Make informed prior $\mathbf{x}_0$ from observed frames $\mathbf{x_1^{0:c}}$
\STATE $\mathbf{x}_\tau = \tau \mathbf{x}_1 + (1-\tau) \mathbf{x}_0$  \hfill \# Interpolate trajectories
\STATE $u \gets \mathbf{x}_1 - \mathbf{x}_0$  \hfill \# Calculate true vector field
\STATE $\mathbf{h} \gets \text{Features}(\mathbf{x}_\tau), \ \mathcal{E} \gets \text{Connectivity}(\mathbf{x}_\tau)$  \hfill \# Construct graphs
\STATE $v_\theta \gets v_\theta(\mathbf{x}_\tau, \mathbf{h}, \mathcal{E}, \tau)$   \hfill \# Forward pass
\STATE $\mathcal{L}_{CFM}(\theta) \gets \| v_\theta - u \|^2$  \hfill \# Calculate loss
\STATE $\theta \gets \theta + \alpha \nabla_\theta \mathcal{L}_{CFM}(\theta)$  \hfill \# Backpropagate and optimize
\ENDWHILE
\end{algorithmic}
\end{algorithm}

\begin{algorithm}

\caption{Inference using Euler Integration}
\label{algo2}
\begin{algorithmic}
\REQUIRE Test Dataset $p_1$, Trained model $v_\theta(\mathbf{x}, \mathbf{h}, \mathcal{E}, \tau)$, Number of Function Evaluations $n$
\FOR{$\mathbf{x}_1 \in p_1$}
\STATE $\mathbf{x}_0^{0:c} \gets \mathbf{x}_1^{0:c}$, \ $\mathbf{x}_0^{c:T} \gets \zeta(\mathbf{x}_1^{0:c})$
 \hfill \# Make informed prior $\mathbf{x}_0$ from observed frames $\mathbf{x_1^{0:c}}$
\STATE $\hat{\mathbf{x}} \gets \mathbf{x}_0$, \ $\Delta \tau \gets \frac{1}{n}$
\FOR{$\tau \in \{0, \frac{1}{n},...,\frac{n-1}{n}\}$}
\STATE $\mathbf{h} \gets \text{Features}(\hat{\mathbf{x}}), \ \mathcal{E} \gets \text{Connectivity}(\hat{\mathbf{x}})$  \hfill \# Construct graphs
\STATE $v_\theta \gets v_\theta(\hat{\mathbf{x}}, \mathbf{h}, \mathcal{E}, \tau)$   \hfill \# Forward pass
\STATE $\hat{\mathbf{x}} \gets \hat{\mathbf{x}} \ + \ \Delta \tau \cdot v_\theta$
 \hfill \# Update $\hat{\mathbf{x}}$ using Euler method
\ENDFOR
\ENDFOR
\end{algorithmic}
\end{algorithm}

\subsection{Definitions of Message Passing, the UNet and the Transformer Architectures}

\label{appendix:mp_layer}

The spatial component of the model updates latent bubble embeddings through message passing over the bubble graph. For an edge \((i,j)\in\mathcal{E}\) at time \(t\), the message from bubble \(j\) to bubble \(i\) is computed as
\begin{equation}
    \mathbf{m}_{ij}^{t}
    =
    \varphi_{\mathbf{m}}
    \left(
    \mathbf{h}_i^{t},
    \mathbf{h}_j^{t},
    \mathbf{e}_{ij}^{t},
    \|\mathbf{x}_i^{t}-\mathbf{x}_j^{t}\|^2,
    \gamma_\tau(\tau)
    \right),
\end{equation}
where \(\mathbf{h}_i^{t}\) and \(\mathbf{h}_j^{t}\) are latent node embeddings, \(\mathbf{e}_{ij}^{t}\) denotes edge features, \(\mathbf{x}_i^{t}-\mathbf{x}_j^{t}\) is the relative displacement, and \(\gamma_\tau(\tau)\) is the projected sinusoidal embedding of the flow-matching time.

Messages are aggregated over neighboring bubbles,
\begin{equation}
    \bar{\mathbf{m}}_i^{t}
    =
    \frac{1}{N}\sum_{j\in\mathcal{N}(i)}
    \mathbf{m}_{ij}^{t},
\end{equation}
and the hidden state is updated with a residual MLP block,
\begin{equation}
    \mathbf{h}_i^{t\,\prime}
    =
    \mathrm{LayerNorm}
    \left(
    \mathbf{h}_i^{t}
    +
    \varphi_{\mathbf{h}}
    \left(
    \mathbf{h}_i^{t},
    \bar{\mathbf{m}}_i^{t},
    \gamma_\tau(\tau)
    \right)
    \right).
\end{equation}
Here, \(\varphi_{\mathbf{m}}\) and \(\varphi_{\mathbf{h}}\) are MLPs with SiLU activations. The updated hidden state $\textbf{h}^\prime$ is used both to update the edge features (via a residual edge MLP, in all but the final layer) and, in parallel, a per-band spherical-harmonic coefficient update through separate output heads. As in the main text, edge features are constructed from relative, not absolute, node quantities, so the layer preserves translational equivariance.

Our temporal convolution layer follows the U-Net design of \cite{brinke2026, dhariwal2021diffusionmodelsbeatgans} with the same parameters. The U-Net takes the latent node embeddings h together with a 16-dimensional sinusoidal embedding of the flow-matching time $\tau$. Since the trajectory data also depends on physical time within the window, we add a sinusoidal frame-index embedding to $\mathbf{h}$. The U-Net applies convolutions along the temporal axis over the T frames of each trajectory. Its output is an updated latent representation $\mathbf{h}$, from which a two-layer $\varphi_{\mathbf{v}}$ predicts the velocity-field update $\Delta \mathbf{v}$, and a separate per-band MLP predicts the spherical-harmonic coefficient update, mirroring the decoding used in the message-passing layer.

The transformer variant substitutes each message-passing (MP) and temporal-convolution (UNet) layer in the main architecture with a single-axis attention block, adapted from LaM-SLidE \cite{lamslide}. As in the main architecture, spatial and temporal processing alternate across depth, each transformer layer instance operates along exactly one axis. Each layer applies scaled dot-product self-attention with rotary positional embeddings (RoPE) computed along its axis: spatial layers attend across the B bubbles at a fixed frame, and temporal layers attend across the T frames of a fixed bubble's trajectory. Query, key, and value projections are computed from a single fused linear layer together with the block's MLP branch, and the attention and MLP outputs are concatenated and mapped back to the hidden dimension by a second linear layer. Query and key vectors are RMS-normalized (QK-norm) prior to attention for training stability. The flow-matching time $\tau$ is embedded via a 256-dimensional sinusoidal embedding and a two-layer MLP, then injected through adaptive layer normalization (AdaLN): each block predicts a per-channel shift, scale, and gate from $\tau$, applied before and after the attention operation, following standard DiT-style conditioning. Node features are projected via a shared input linear layer plus a separate conditioning-embedding path, summed before entering the attention stack. Only the temporal blocks are followed by output decoder heads: a two-layer MLP predicts the positional velocity update, and one MLP per spherical-harmonic band $\ell$ predicts the corresponding coefficient update, matching the per-band decoding used elsewhere in the architecture. Spatial blocks update only the latent node embedding and do not produce a direct output at that layer.

\subsection{Hyperparameters} \label{appendix:hyperparam}
The hyperparameters used to train and evaluate the model are included in Table~\ref{appendix:hyper}. We use AdamW with learning rate $5\cdot10^{-4}$ and otherwise default parameters and a ReduceLROnPlateau scheduler based on the minimum validation loss, with a decay factor of 0.5 and patience of 50 epochs. Graph connectivity uses a cutoff of 20mm, and we used a batch size of 4.  SH coefficients $\mathbf{c}$ are min-max normalized to $[-3,3]$.  The deterministic setup uses a fixed prior with lower epochs, has very small positional and SH coefficient noise added during training and dropout to reduce overfitting. The transformer experiment uses more hidden dimensions, 1 layer deeper to close the gap of number of model parameters with the other configurations and positions $\mathbf{p}$ as additional node features as it lacks edge features to get a sense of the geometry of the trajectories. 

\begin{table}[h]
    \centering
    \caption{Training hyperparameters for the experiments.}
    \begin{small}
    \setlength{\tabcolsep}{2pt}
    \begin{tabular}{lcccccccccc}
    \toprule
    \textbf{Model} 
    & \textbf{SH order} 
    & \textbf{Prior $\mu$}
    & \textbf{Prior $\sigma_\mathbf{p},\sigma_\mathbf{c}$}
    & \textbf{Prior $s$}
    & \textbf{Layers} 
    & \textbf{Epochs} 
    & \textbf{Hidden dim}
    & \textbf{Dropout} \\
    
    \midrule
    Model \((L=5)\) 
    & 5 
    & $\langle\Delta \mathbf{x}_{0:T_c}\rangle$
    & $\sigma_{0:T_c}\cdot s$
    & $1$
    & $3 \times$ (MP,U-Net) 
    & 400
    & 64
    & 0 \\

    Gaussian prior
    & 5 
    & $\mathbf{x}_{T_c}$
    & $0.01, 0.05$
    & -
    & $3 \times$ (MP,U-Net) 
    & 400
    & 64
    & 0\\

    Deterministic
    & 5 
    & $\langle\Delta \mathbf{x}_{0:T_c}\rangle$
    & 0
    & -
    & $3 \times$ (MP,U-Net) 
    & 100
    & 64
    & 0.05 \\

    Independent 
    & 5 
    & $\langle\Delta \mathbf{x}_{0:T_c}\rangle$
    & $\sigma_{0:T_c}\cdot s$
    & 1
    & $3 \times$ U-Net 
    & 400
    & 64
    & 0\\
    
    Transformer
    & 5 
    & $\langle\Delta \mathbf{x}_{0:T_c}\rangle$
    & $\sigma_{0:T_c}\cdot s$
    & 1
    & $4 \times$ (Space,Time) 
    & 400
    & 128
    & 0  \\

    Model \((L=3)\) 
    & 3 
    & $\langle\Delta \mathbf{x}_{0:T_c}\rangle$
    & $\sigma_{0:T_c}\cdot s$
    & 1
    & $3 \times$ (MP,U-Net) 
    & 400
    & 64
    & 0 \\

    \bottomrule
    \end{tabular}
    \end{small}
    \label{appendix:hyper}
\end{table}

\subsection{Compute resources} \label{appendix:res}
All model experiments were run on either a single NVIDIA RTX 3080 or Tesla V100 GPU and need around 4-8GB of video memory. The model configurations required approximately 2-8 hours of training depending on the spherical harmonic order and whether message passing or a transformer was used. Inference was comparatively inexpensive and took only a few minutes per experiment, evaluation takes 10-30 minutes because of the many individual point clouds that need to be compared.



\section{Evaluation} \label{app:eval}

\subsection{Evaluation Details}
\label{app:eval-details}

This section gives precise definitions of every quantity reported in Table \ref{results}.

\textbf{Positional errors}
For a predicted position $\hat{p}_i(t)$ and its ground truth $p_i(t)$, we define the (unnormalized)
average and final displacement errors over the prediction window as
\begin{align}
    \mathrm{ADE} &= \Big\langle \, \|\hat{p}_i(t) - p_i(t)\| \, \Big\rangle_{t \in T_c+1:T} \\
    \mathrm{FDE} &= \big\langle \, \|\hat{p}_i(T) - p_i(T)\| \, \big\rangle
\end{align}
where $\langle \cdot \rangle$ denotes the mean over bubbles and samples. The relative variants
reported in Table \ref{results} normalize by the ground-truth arc length accumulated \emph{within the
prediction window},
\begin{equation}
    \ell_i = \sum_{t=T_c+1}^{T} \big\| p_i(t) - p_i(t-1) \big\|,
\end{equation}
giving
\begin{equation}
    \mathrm{R\text{-}ADE} = \Big\langle \, \|\hat{p}_i(t) - p_i(t)\| \, / \, \ell_i \, \Big\rangle,
    \qquad
    \mathrm{R\text{-}FDE} = \big\langle \, \|\hat{p}_i(T) - p_i(T)\| \, / \, \ell_i \, \big\rangle.
\end{equation}

\textbf{Deformation errors}
At a subset of evaluation frames, we reconstruct a point cloud on the bubble surface from
the spherical-harmonic coefficients $\mathbf{c}$ and take its convex hull for volume and surface area calculations. Given predicted and
ground-truth hulls $\hat{\mathcal{H}}_i(t)$, $\mathcal{H}_i(t)$ at frame $t$, we report
\begin{equation}
    \mathrm{IoU} = \Big\langle \, \frac{\mathrm{vol}\big(\hat{\mathcal{H}}_i(t) \cap \mathcal{H}_i(t)\big)}
    {\mathrm{vol}\big(\hat{\mathcal{H}}_i(t) \cup \mathcal{H}_i(t)\big)} \, \Big\rangle,
\end{equation}
and the mean Chamfer distance between the corresponding surface point clouds $\hat{P}_i(t)$,
$P_i(t)$,
\begin{equation}
    \mathrm{CD} = \big\langle \, CD\big(\hat{P}_i(t), P_i(t)\big) \, \big\rangle,
    \ 
    CD(A, B) = \frac{1}{|A|}\sum_{a \in A} \min_{b \in B} \|a-b\|
    + \frac{1}{|B|}\sum_{b \in B} \min_{a \in A} \|a - b\|.
\end{equation}

\textbf{Distributional alignment}
For two empirical marginal distributions $P$, $Q$ of a scalar quantity, we define the
IQR-normalized Wasserstein-1 distance
\begin{equation}
    \mathcal{W}_{IQR}(P, Q) = \frac{W_1(P, Q)}{\tfrac{1}{2}\big(\mathrm{IQR}(P) + \mathrm{IQR}(Q)\big) + \varepsilon},
\end{equation}
where $W_1$ is the standard 1-Wasserstein distance, $\mathrm{IQR}(\cdot)$ denotes the
interquartile range (75th minus 25th percentile), and $\varepsilon = 10^{-8}$ avoids division by
zero. Each of the three columns in Table \ref{results} averages $\mathcal{W}_1^{\text{norm}}$ over a small
set of constituent quantities:
\begin{itemize}
    \item \textbf{Kinematics} (mean of 3): velocity magnitude $\|\dot{p}_i(t)\|$ (normalized by
    bubble diameter), rise angle $\arccos\!\big(\dot{p}_{i,z}(t) / \|\dot{p}_i(t)\|\big)$, and
    trajectory tortuosity, i.e.\ the ratio of arc length to end-to-end chord length over the
    prediction window.
    \item \textbf{Morphology} (mean of 3): convex-hull surface area (normalized by squared bubble
    diameter), sphericity $\Psi = \pi^{1/3}(6V)^{2/3} / A$ computed from convex-hull volume $V$
    and surface area $A$, and surface deformation velocity, the mean absolute time-derivative of
    each surface point's radial distance from the bubble centroid (normalized by bubble diameter).
    \item \textbf{Interaction} (mean of 2): nearest-neighbor distance between bubble centroids
    (normalized by bubble diameter, thresholded below $1.4$ diameters), and close-neighbor count,
    the number of neighboring bubbles within $2$ diameters of a given bubble's centroid.
\end{itemize}

\textbf{Reported scores}
Table \ref{results} reports $\mathrm{R\text{-}ADE}$, $\mathrm{R\text{-}FDE}$, $\mathrm{IoU}$,
$\mathrm{CD}$, and the three grouped $\mathcal{W}_{IQR}$ scores above,
labeled Kinematics, Morphology, and Interaction.

\subsection{Additional Analysis}

\begin{figure}[h!]
    \centering
    \includegraphics[width=0.95\linewidth]{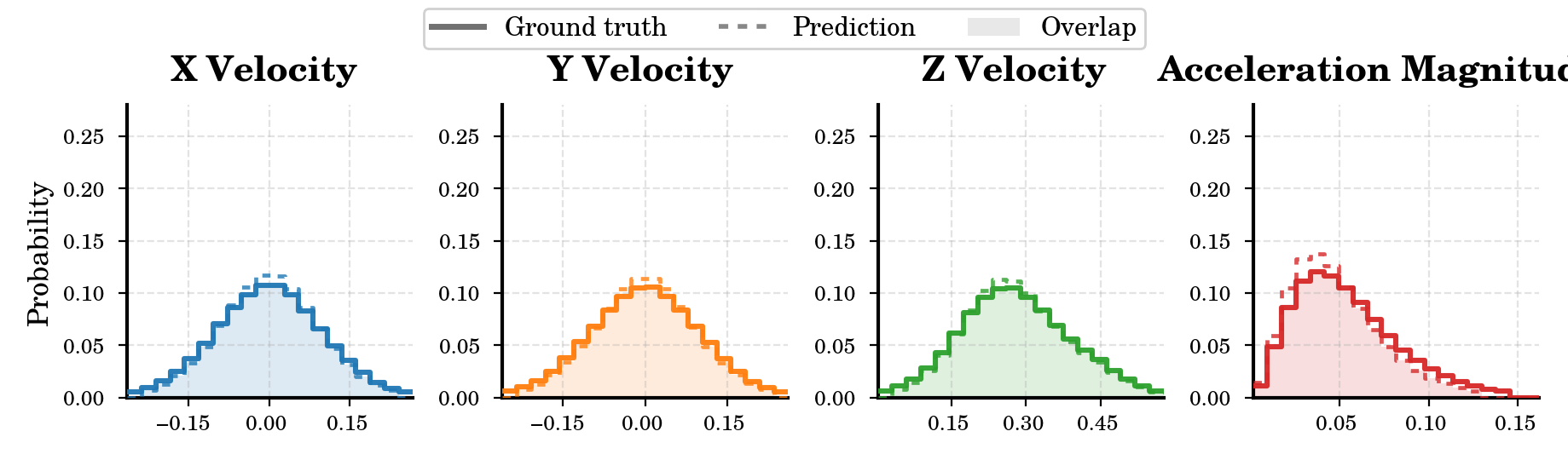}
    \caption{Predicted and ground truth probability mass functions of position-based marginal quantities from the $T_c+1:T$ frames, from the main experiment \hyperref[results]{\textit{(iii)}}.}
    \label{fig:densities2}
\end{figure}

\begin{figure}[h!]
    \centering
    \includegraphics[width=0.95\linewidth]{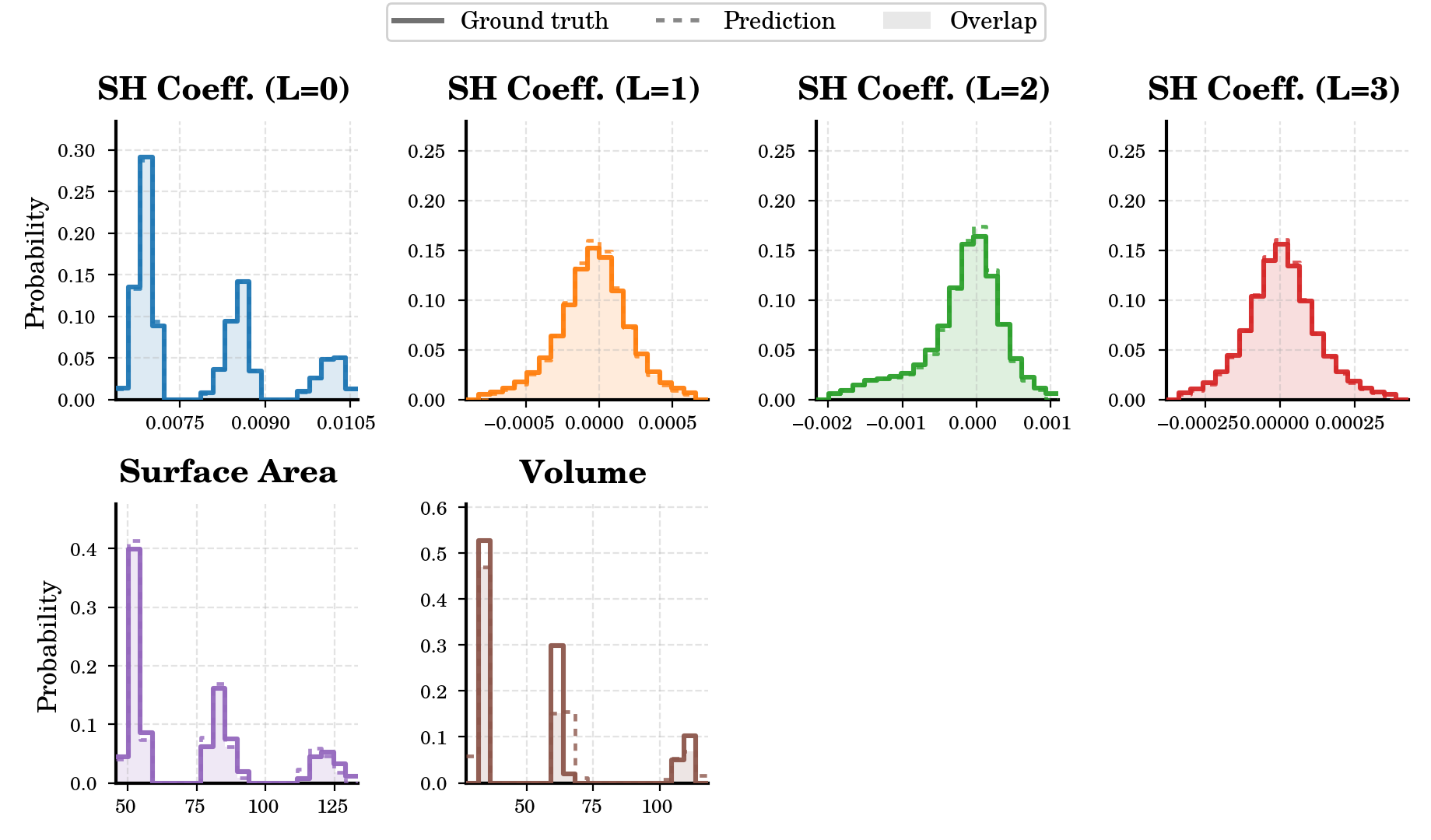}
    \caption{Predicted and ground truth probability mass functions of shape-based marginal quantities. From the main experiment \hyperref[results]{\textit{(iii)}}.}
    \label{fig:densities3}
\end{figure}

We plot the predicted and ground truth empirical probability mass functions of x,y and z velocities, acceleration magnitude, SH coefficients of $L=0$ up to $L=3$, surface area and volume from the main experiment \hyperref[results]{\textit{(iii)}} in Figures \ref{fig:densities2} and \ref{fig:densities3}. The model faithfully reproduces all distributions, but shows slight bias in the acceleration magnitudes.

\begin{figure}[h!]
    \centering
    \includegraphics[width=1\linewidth]{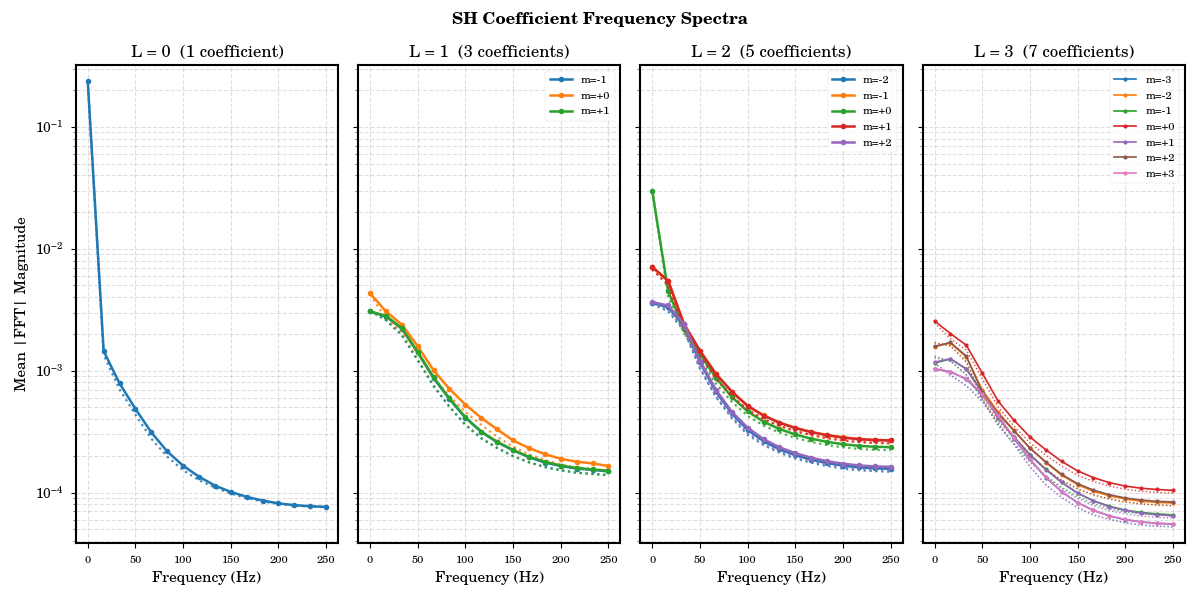}
    \caption{Frequency analysis of the spherical harmonic coefficients trajectories $\mathbf{c}$ at frames $T_c:T$. The solid lines are ground truth, the predictions are dotted. From the main experiment \hyperref[results]{\textit{(iii)}}.}
    \label{fig:fourier}
\end{figure}

A spectral analysis of the spherical harmonics coefficient trajectories is shown in Figure \ref{fig:fourier}. The model seems to be slightly biased towards predicting lower-magnitude signals across all frequencies. We see that the $L=0$ coefficient, which represent the bubble volume, has the highest magnitude, which is expected. Notably, the $L=2, m=0$ coefficient, which expresses a vertical squeezing deformation, has the second highest magnitude, as many bubbles are shaped more like an ellipsoid. 

\newpage
\section{Visualizations} \label{app:viz}
\begin{figure}[h!]
    \centering
    \includegraphics[width=1\linewidth]{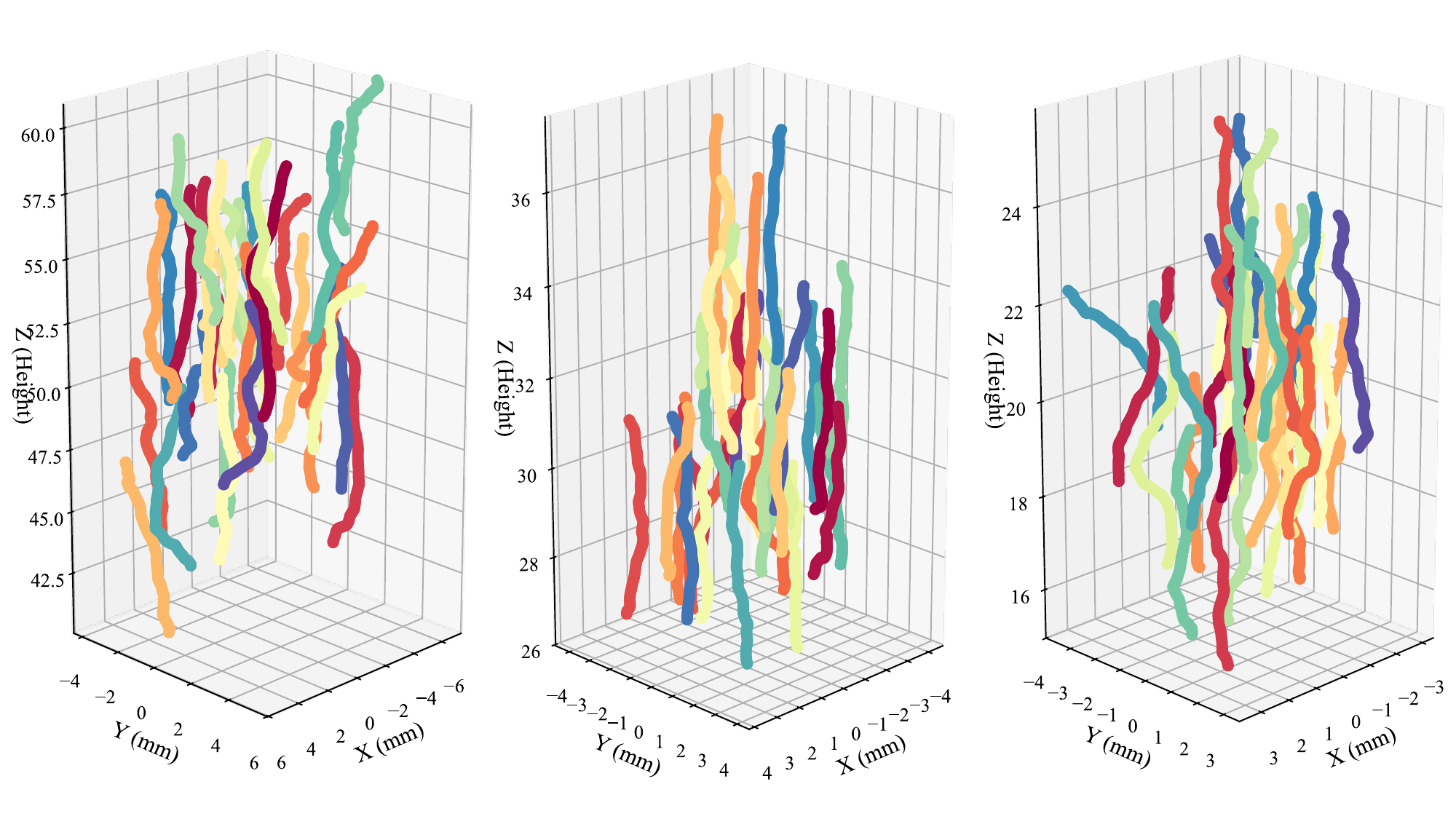}
    \caption{Visualizations of centered 32 bubble centroid trajectories. Duration is 200 ms, number of timesteps is 100, filtered to exclude periodic boundary jumps.}
    \label{fig:centroid_viz}
\end{figure}

\begin{figure}[h!]
    \centering
    \includegraphics[width=0.8\linewidth]{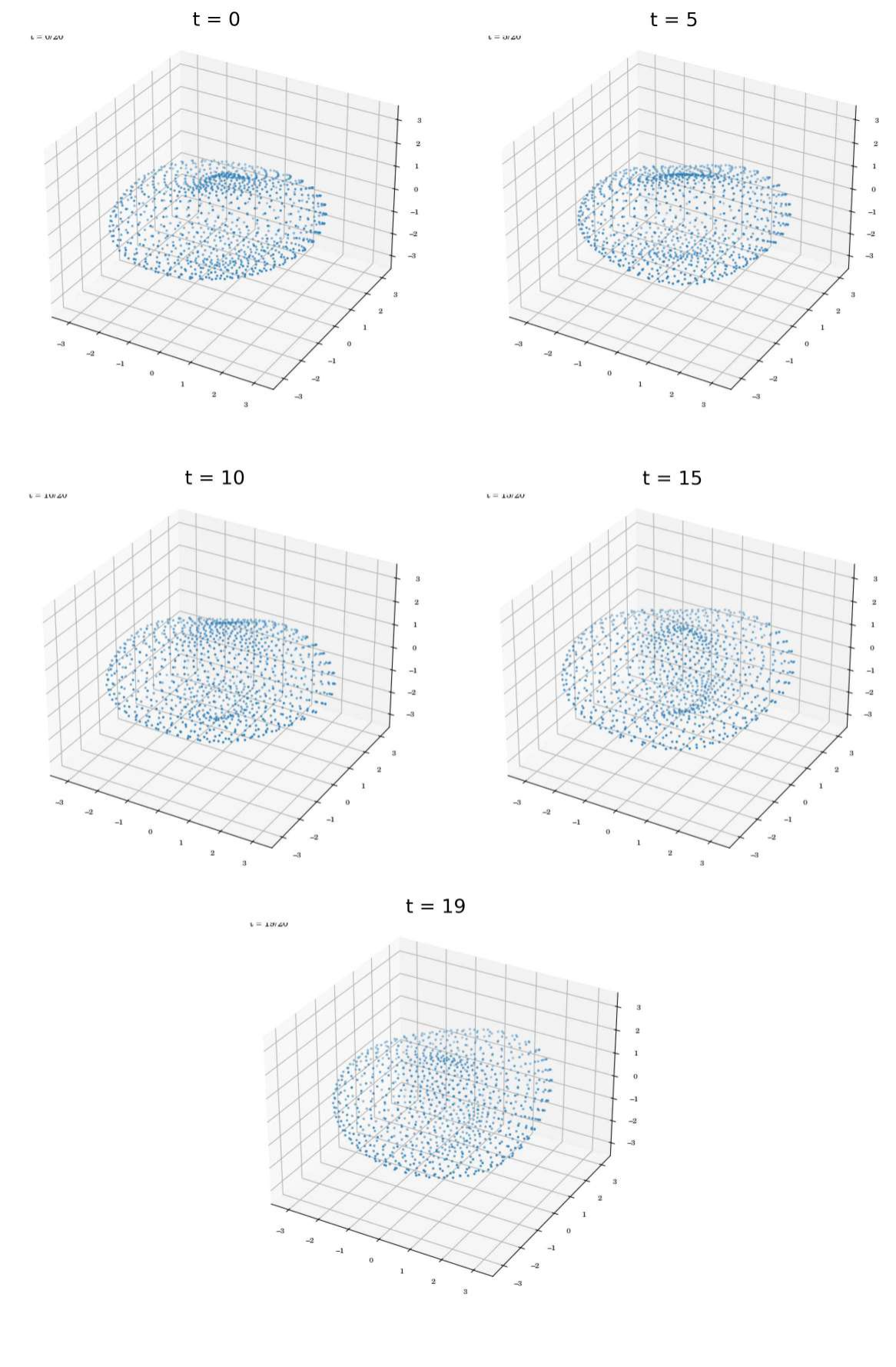}
    \caption{Visualizations of the shape of one bubble changing over time, taken from \cite{ramesh2026bubbleshdatasetrisingbubbles}.}
    \label{fig:points}
\end{figure}

We include more visualizations of the data in Figure \ref{fig:centroid_viz} and Figure \ref{fig:points}, taken from \cite{ramesh2026bubbleshdatasetrisingbubbles}.

\vspace{-3em}

\end{document}